\documentclass[sn-mathphys-num]{sn-jnl}

\usepackage{graphicx}%
\usepackage{float}%
\usepackage{color}
\usepackage{multirow}%
\usepackage{amsmath,amssymb,amsfonts,csquotes}%
\usepackage{amsthm}%
\usepackage{mathrsfs}%
\usepackage[title]{appendix}%
\usepackage{xcolor}%
\usepackage{textcomp}%
\usepackage{manyfoot}%
\usepackage{booktabs}%
\usepackage{algorithm}%
\usepackage{algorithmicx}%
\usepackage{algpseudocode}%
\usepackage{listings}%
\usepackage{bigstrut,multirow,rotating,booktabs,graphicx}

\theoremstyle{thmstyleone}%
\theoremstyle{thmstyletwo}%

\theoremstyle{thmstylethree}%

\begin{document}

\title[Article Title]{
HeGraphAdapter: 
Tuning Multi-Modal Vision-Language Models with
Heterogeneous Graph Adapter
}

\author[]{\sur{Yumiao Zhao}}

\author*[]{\sur{Bo Jiang}*}\email{jiangbo@ahu.edu.cn}

\author[]{\sur{Xiao Wang}}

\author[]{\sur{Qin Xu}}

\author[]{\sur{Jin Tang}}


\affil []{\orgdiv{School of Computer Science and Technology}, \orgname{Anhui University}}



\abstract{Adapter-based tuning methods have shown significant potential in transferring knowledge from pre-trained Vision-Language Models (VLMs) to the downstream tasks. 
However, after reviewing existing adapters, we find they generally fail to \emph{fully} explore the interactions between different modalities in constructing task-specific knowledge. Also, existing works usually only focus on similarity matching (or relation mining) between positive text prompts, making it challenging to distinguish the classes with high similar visual contents. 
To address these issues, in this paper, we propose a novel Heterogeneous Graph Adapter (HeGraphAdapter) to achieve tuning VLMs for the downstream tasks. 
To be specific, we first construct a unified heterogeneous graph mode, which contains i) visual nodes, positive text nodes and negative text nodes, and ii) several types of edge connections to comprehensively model the intra-modality, inter-modality and inter-class structure knowledge together. Next, we employ a specific Heterogeneous Graph Neural Network (HGNN) to excavate multi-modality (heterogeneous) structure knowledge for adapting both visual and textual features for the downstream tasks. Finally, after HeGraphAdapter, we construct both text-based and visual-based classifiers simultaneously to comprehensively enhance the performance of the CLIP model.
Experimental results on 11 benchmark datasets demonstrate the effectiveness and benefits of the proposed HeGraphAdapter. 
\textcolor{blue}{The code will be released upon acceptance.} }





\maketitle
\section{Introduction}\label{sec1}
Recently, large-scale Vision-Language Models (VLMs), such as CLIP~\cite{radford2021learning} and its variants~\cite{jia2021scaling,zhang2024long,li2024llava}, provide a new paradigm for computer vision and demonstrate superior generalization performance on many downstream tasks, e.g., zero-shot generalization~\cite{smith2023continual}, few-shot classification~\cite{zhou2022learning}, and out-of-distribution (OOD) detection~\cite{nie2024out}. VLMs are generally pre-trained on millions of paired visual-language data, enabling them to understand open-vocabulary visual concepts. They demonstrate excellent generalization ability with hand-crafted text prompts like ``A photo of a \{class\}''.  Overall, there exist three kinds of transfer learning methods to transfer the knowledge of VLMs to the downstream tasks, i.e., full fine-tuning, prompt tuning and adapter-style tuning. Full fine-tuning methods~\cite{lv2023full,wu2023yuan} focus on adjusting all parameters of the pre-trained model to adapt it for specific downstream tasks. However, these methods require large resources during the fine-tuning process and also face the risk of over-fitting in low-data scenarios. 
Prompt-tuning~\cite{,gao2024clip,zhou2022learning,khattak2023maple} methods aim to introduce learnable prompts on the pre-trained model to adjust the features for the downstream tasks. For example, some works leverage hand-crafted templates like ``A photo of a \{class\}'' or design learnable text prompts to fine-tune the CLIP model. However, these methods require computing the gradients of the text encoder during training, usually leading to high computational cost. 
Adapter-style tuning methods~\cite{gao2024clip,udandarao2023sus,yu2023task,zhu2023not,tang2024amu,caron2021emerging} focus on refining the textual or visual features on the output side to improve the task-specific performance. 
For instance, TaskRes~\cite{huang2022task} introduces residual vectors with textual features to learn task-specific knowledge. 
Tip-Adapter~\cite{zhang2022tip} constructs a key-value cache model from visual features to refine the classifier to the downstream tasks. 
AMU-Tuning~\cite{tang2024amu} introduces a visual auxiliary modal to refine the classifier to improve the performance of the downstream tasks.
However, these adapter-style approaches have two main limitations: 1) only use a single modality to learn task-specific knowledge, and 2) fail to exploit the inherent structural relationships between textual and visual features in the downstream tasks. 

To overcome the above limitations, GraphAdapter~\cite{li2024graphadapter} has been proposed recently to learn task-specific knowledge by leveraging a dual knowledge graph learning model. In particular, GraphAdapter~\cite{li2024graphadapter} first adopts both textual and visual knowledge graphs to model the inter-class relationship for textual and visual modality respectively and employs a regular graph convolutional network~\cite{kipf2016semi} to excavate the knowledge for adapting the textual features in both graphs. Then, it fuses the multi-modality graph node representations via the weighted `sum' fusion strategy. 
Despite that, we observe that GraphAdapter has three main shortcomings: 
i) it combines multi-modality knowledge via the simple weighted `sum' fusion strategy, which fails to fully explore the interactions between different modalities. 
ii) It only considers the positive prompts for text nodes, making it challenging to distinguish the classes with high visual similarity. 
iii) It only refines the text modality features by exploiting dual graphs, overlooking the adjustment of the visual modality features to determine the final classifier. 

To address these problems, in this paper, we propose a novel Heterogeneous Graph Adapter (HeGraphAdapter) for tuning VLMs. 
In contrast to the previous adapter methods~\cite{li2024graphadapter,huang2022task,zhang2022tip}, our proposed HeGraphAdapter leverages a \textbf{heterogeneous graph learning} to fully capture the multi-modality knowledge cooperatively to adapt the textual and visual features for the downstream tasks. Consequently, we first propose a heterogeneous graph that involves three types of nodes, i.e., visual nodes, positive text nodes and negative text nodes to denote the `class' representations under different modalities respectively.  
The positive and negative text nodes are initialized by using the CLIP features of the positive prompt (``A photo of \{class\}") and negative prompt (``A photo of \emph{no} \{class\}") respectively, 
while the visual nodes are initialized by computing the `mean' features of visual images within each class. Also, there are six types of edge connections in our heterogeneous graph which encode the rich relations between different types of nodes, as presented in Section~\ref{construction} in detail. 
Then, we develop a specific Heterogeneous Graph Neural Network (HGNN) to excavate the heterogeneous structure knowledge and obtain task-specific textual and visual features. Finally, we construct both text-based and visual-based classifiers simultaneously to comprehensively enhance the performance of CLIP model for the downstream tasks. 
Overall, the main contributions of this paper are summarized as follows:
\begin{itemize}
  \item We propose a novel HeGraphAdapter for tuning multi-modality VLMs. It can comprehensively model the intra-modality, inter-modality, and inter-class structure knowledge simultaneously in a unified model. 
  To the best of our knowledge, this work is the first study to exploit the {heterogeneous graph learning} for multi-modality VLM fine-tuning problem. 
  
  \item To obtain the distinctive task-specific knowledge for the downstream tasks, we introduce negative text prompts into our heterogeneous graph learning to fully model both similarity and dissimilarity relationships among different class nodes. 
  
  \item We construct both text-based and visual-based classifiers simultaneously to comprehensively enhance the performance of CLIP for the downstream tasks.

\end{itemize}
Extensive experiments on eleven benchmark datasets demonstrate the superior performance of our HeGraphAdapter on various few-shot classification tasks when compared with many other related approaches.  


\section{Related Work}\label{sec2}

\textbf{Tuning for vision-language model: } Vision-language models (VLMs) aim to learn the visual and language representation simultaneously, through pre-training on large-scale text-image pair datasets. Extensive studies have evaluated that pre-trained VLMs possess remarkable capabilities in zero-shot learning~\cite{nag2022zero,radford2021learning}, few-shot classification~\cite{zhou2022learning,yu2023task}, and cross-modality generation tasks~\cite{nichol2021glide,ramesh2022hierarchical}. Existing works focus on exploring lightweight fine-tuning techniques to transfer the knowledge of VLMs and adapt them to downstream tasks. There are three categories of fine-tuning techniques: full fine-tuning, prompt tuning, and adapter-style tuning. Full fine-tuning methods~\cite{lv2023full,wu2023yuan} adjust all parameters of the pre-trained model using task-specific data. LOMO~\cite{lv2023full} design a low-memory optimization method to reduce memory usage during fine-tuning by fusing the gradient computation and the parameter update. 
Prompt tuning methods~\cite{zhou2022conditional,zhou2022learning,khattak2023maple,xing2023dual} are inspired by the natural language process, which focuses on introducing learnable prompts on the pre-trained model for adapting the downstream tasks. For example, Zero-shot CLIP~\cite{radford2021learning} uses a hand-crafted template, like ``A photo of a \{class\}", and reports superior generalization performance in the downstream task. CoOp~\cite{zhou2022learning}
introduce a learnable prompt for the text-encoder to learn the task-specific knowledge. CoCoOp~\cite{zhou2022conditional} designs a lightweight meta-network to generate text prompts for each image and enhances the generalization ability of CoOp. MaPLe~\cite{khattak2023maple} introduces branch-aware hierarchical prompts to fine-tune the text and image encoder representations simultaneously.
Adapter-style tuning methods~\cite{gao2024clip,udandarao2023sus,yu2023task,zhu2023not,tang2024amu,caron2021emerging}  adapt to the downstream tasks by turning the feature representations of visual or textual on the output side. CLIP-Adapter~\cite{gao2024clip} design an MLP-based adapter module to fine-tune the textual and visual features on the output side. 
AMU-Tuning~\cite{tang2024amu} learns an effective logit bias to improve the performance for the downstream tasks by introducing auxiliary features generated by other pre-trained large models. GraphAdapter~\cite{li2024graphadapter} constructs a dual knowledge graph to model the inter-class relationships for adapting the textual features for the downstream tasks. 

\textbf{Heterogeneous graph learning: } Heterogeneous graph (HeGraph) representation learning~\cite{sun2013mining,wang2022survey, zeng2024multi} performs excellent capability in formalizing the structural knowledge between different types of nodes and relations, which has been widely applied in real-world scenarios, like recommendation systems~\cite{shi2018heterogeneous}, healthcare systems~\cite{li2024gtp}, and computer vision~\cite{wang2024integrated}. The core challenge for HeGraph is constructing a graph that needs to consider the difference in neighbor information under different relations and realize hierarchical aggregation. There are two main categories of Heterogeneous Graph Neural Networks (HGNN): Meta-path-based~\cite{fu2020magnn,yang2023simple} and Meta-path-free~\cite{fu2020magnn,hong2020attention}. The Meta-path-based methods first aggregate the structural information from meta-paths with the same semantics, then fuse the information from different meta-paths with different semantics. In contrast, the Meta-path-free methods integrate the structural and semantic information simultaneously.

\section{Approach}\label{sec3}

\begin{figure*}[h]
  \centering
  \includegraphics[width=\linewidth]{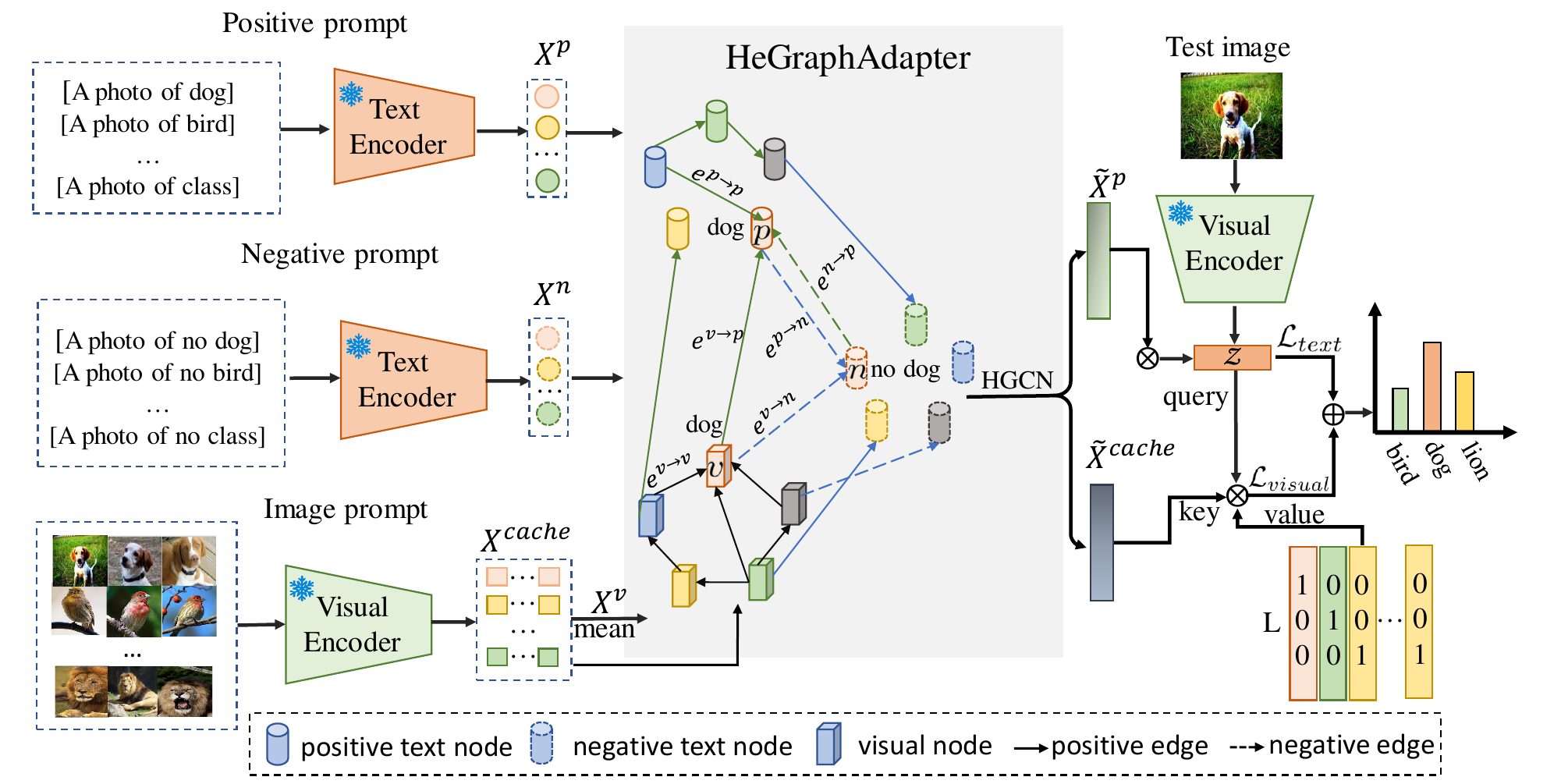}
  \caption{ The pipeline of the proposed method mainly contains prompt generation, HeGraphAdapter and label prediction.} 
\label{backbone_v}
\end{figure*}

In this section, 
we present our Heterogeneous Graph Adapter (HeGraphAdapter) that comprehensively models the intra-modality, inter-modality and inter-class structure knowledge simultaneously in a unified model by considering positive text prompts, negative text prompts and visual prompts together. 
The whole framework of the proposed VLM tuning method is depicted in Fig.~\ref{backbone_v}, which involves prompts generation, HeGraphAdapter and label prediction.

\subsection{Heterogeneous Graph Adapter}

\subsubsection{Heterogeneous graph construction}\label{construction}
To excavate the structure knowledge for the downstream tasks and obtain a discriminative classifier, we construct a heterogeneous graph $\mathcal{G}=(\mathcal{V},\mathcal{E}, \mathcal{X}, \mathcal{R})$ to capture the relationships between different classes
where $\mathcal{V}=\left\{V^p, V^n, V^v\right\}$ denotes the node set containing three kinds of nodes and $\mathcal{E}=\{E^{n{\rightarrow}p}, E^{v{\rightarrow}p}, E^{p{\rightarrow}p}, E^{p{\rightarrow}n}, E^{v{\rightarrow}n}, E^{v{\rightarrow}v}\}$ denotes the edge set including six kinds of edges. $\mathcal{X}$ and $\mathcal{R}$ denotes the collection for node and edge features respectively. The details of them are introduced below. 

\textbf{Nodes.} For the nodes $\mathcal{V}$, it involves three types of nodes, i.e., positive text nodes $V^p$, negative text nodes $V^n$, and visual nodes $V^v$ to denote the `class' representation of different modalities respectively. 
We use $\mathcal{X} = \{X^p, X^n, X^v\}$ to represent the collection of features for positive text nodes, negative text nodes and visual nodes. Given one specific task with a $C$-class fine-tuning dataset, we design some positive text prompts for each class (e.g., ``A photo of a \{class\}”) and encode it with CLIP modal~\cite{radford2021learning} to initialize the feature representations of positive text nodes $X^p=\{x_1^p,x_2^p,\cdots,x_C^p\}$, $x_i^p\in \mathbb{R}^{d}$, as suggested in work~\cite{li2024graphadapter}. 
Also, to excavate the negative semantic knowledge of each class, we also introduce negative text prompts for each class (e.g., ``A photo of \textbf{no} \{class\}”)~\cite{wang2023clipn} and encode it with CLIP modal to initialize the features of negative text nodes $X^n=\{x_1^n,x_2^n,\cdots,x_C^n\}$. In addition, we use the visual encoder of CLIP to obtain the visual features of all fine-tuning images $X^{cache}$ and compute the `mean' feature within each class to initialize the features of visual nodes as $X^v=\{x_1^v,x_2^v,\cdots, x_C^v\}$. 

\textbf{Edges.} As shown in Fig.~\ref{backbone_v}, for the edges $\mathcal{E}$, it involves six types of edge connections,
where $E^{n{\rightarrow}p}$ denotes the directed edges from negative text nodes to positive text nodes, $E^{v{\rightarrow}p}$, edges from visual nodes to positive text nodes, $E^{p{\rightarrow}p}$, edges between positive text nodes, $E^{p{\rightarrow}n}$, edges from positive text nodes to negative text nodes, $E^{v{\rightarrow}n}$, edges from visual nodes to negative text nodes, and $E^{v{\rightarrow}v}$ denote the edges between visual nodes. 
Let $\mathcal{R}=\{\cdots, R^{p{\rightarrow}p}, \cdots\}$ denote the associated weight value for each edge in $\mathcal{E}$, i.e., $R^{p{\rightarrow}p}=\{r_{ij}^{p{\rightarrow}p}\}^C_{i,j=1}$, and similarly defined for other type edges. 
The detailed computations for these weights are introduced below. 
First, for the edges connecting positive text and visual nodes, i.e., $r_{ij}^{p{\rightarrow}p}, r_{ij}^{v{\rightarrow}p}$ and $r_{ij}^{v{\rightarrow}v}$, we compute the edge weight as the cosine similarity between node's features. 
Then, for the edges existing from each positive text/visual node to its corresponding negative text node, i.e.,  $r_{ii}^{p{\rightarrow}n}, r_{ii}^{v{\rightarrow}n}$ and $r_{ii}^{n{\rightarrow}p}$. We compute the edge weight as the dissimilarity between node features, e.g., $r_{ii}^{p{\rightarrow}n}=- \frac{x_i^p x_i^n} {\left \|x_i^p  \right \| \left \|  x_i^n\right \| }$, $r_{ii}^{v{\rightarrow}n}=- \frac{x_i^v x_i^n} {\left \|x_i^v  \right \| \left \|  x_i^n\right \|}$, because the negative nodes encode the opposite information of the positive nodes. 
Third, 
for the edges existing from each positive text/visual node to the other negative text nodes, i.e., $r_{ij,j\ne i}^{p{\rightarrow}n}$ and $r_{ij,j\ne i}^{v{\rightarrow}n}$, 
we simply compute the edge weight as the cosine similarity between node features.

\subsubsection{Heterogeneous graph message passing}\label{HGNN}
After constructing the heterogeneous graph, we then leverage a Heterogeneous Graph Neural Network (HGCN), as inspired by works~\cite{yang2023simple,wang2019heterogeneous,zhang2019heterogeneous}, to conduct message passing across different types of nodes and relationships. Here, we develop a knowledge-guided hierarchical aggregation which consists of three steps: negative text node aggregation, positive text node aggregation and visual node aggregation, as introduced below. 
In the following, we denote 
$X=(x_1,x_2\cdots x_{3C})=(X^p,X^n,X^v)=(x^p_1\cdots x^p_C,x^n_1\cdots x^n_C,x^v_1\cdots x^v_C)$ and use them alternatively.  


\textbf{Negative text node aggregation:}
Hand-crafted negative text prompt often fails to understand ``no'' effectively in the downstream tasks. 
Therefore, we allow the negative text nodes to obtain the contextual knowledge information from their neighboring visual and positive text nodes. 
To be specific, we consider the heterogeneous semantic relationships within the meta-paths $\Phi^n=\left\{ p{\rightarrow}n,  v{\rightarrow}n\right \}$ and aggregate the features from neighbors based on the meta-paths $\Phi^n$. The aggregated feature of the $i$-th negative text node is calculated as follows:
\begin{align}
\label{SignGCN}
&m_i^{\phi}= \sigma \Big( \frac{1}{d_i+1}W^{n}x_i+\sum_{j\in N_i^{\phi}} \frac{r_{ij}^{\phi}}{\sqrt{(d_i+1)(d_j+1) } }W^{n}x_j  \Big)\\
&\tilde{x}_i^n{\longleftarrow}x_i^{n}+\sum_{\phi \in \Phi^{n}}\beta^{\phi} m_i^{\phi}
\end{align}
where $\phi\in \Phi^n$ and 
 $W^{n}$ is a learnable projection matrix.
 $N_i^{\phi}$ denotes the neighbors connected to the node $i$ based on each meta-path $\phi$ and $d_i= {\textstyle \sum_{j\in N^{\phi}_i}
 |r_{ij}^{\phi}  |} $. 
$\beta^\phi$ denotes the weight of the meta-path $\phi$. 
After negative node aggregation, we renew $X^n$ to $\tilde{X}^n$ and 
update the current feature matrix 
$X$ for all nodes as $\tilde{X}=(X^p, \tilde{X}^n, X^v)$ for the following process.

\textbf{Positive text node aggregation:}
To obtain the discriminative textual feature of each class, positive text nodes can aggregate the information from negative text nodes and visual nodes. The heterogeneous relationships of positive text nodes are customized as $\Phi^p=\left\{ p{\rightarrow}p, v{\rightarrow}p, n{\rightarrow}p\right \}$. 
We aggregate the neighbors of each positive text node as follows: 
\begin{align}
\label{SignGCN2}
& m_i^\phi= \sigma \Big( \frac{1}{d_i+1}W^p\tilde{x}_i+\sum_{j\in N_i^{\phi}} \frac{r_{i,j}^{\phi}}{\sqrt{(d_i+1)(d_j+1) } }W^p\tilde{x}_j\Big)\\
\label{alpha}
&\tilde{x}_i^{p}{\longleftarrow}x_i^{p}+\sum_{\phi \in \Phi^{p}}\alpha^{\phi} m_i^{\phi}
\end{align}
where $\phi\in \Phi^P$ and $W^p$ denotes a learnable transformation matrix. $N_i^{\phi}$ denotes the neighbors connected to the $i$-th positive text node based on meta-path $\phi$. $\alpha^\phi$ represents the weight of the meta-path $\phi$. After positive text node aggregation, we update the features of positive text prompts $X^p$ to $\tilde{X}^p$. The feature matrix of all nodes is renewed as $\tilde{X}=(\tilde{X}^p, \tilde{X}^n, X^v)$.

\textbf{Visual node aggregation:}
To learn the visual bias for visual nodes and refine the visual features $X^{cache}$ of all fine-tuning samples, we also consider exploiting the structure information of visual nodes for contextual learning. 
Specifically, we learn the visual bias of the $i$-th visual node by aggregating the information of its neighbors as: 
\begin{align}
\label{SignGCN3}
&m_i^{bias}= \sigma \Big( \sum_{j\in N_i} \frac{r_{i,j}^{v{\rightarrow}v}}{\sqrt{(d_i+1)(d_j+1) } }W^v x^{v}_j \Big)\\
&\tilde{x}_i^{cache}{\longleftarrow}x_i^{cache}+\gamma \cdot m_i^{bias}
\end{align}
where $W^v$ is a learnable transformation matrix and $N_i$ denotes the neighbors connecting to the visual node $i$ based on meta-path $\Phi^v=\{v\rightarrow v\}$. The weight $\gamma$ is a hyper-parameter.

\subsection{Tuning for Downstream Tasks}\label{tuning}
After HeGraphAdapter, we can obtain task-specific textual features as $\widetilde{X}^p$, and visual features $\widetilde{X}^{cache}$ for all fine-tuning samples. Then, we construct both text-based and visual-based classifiers simultaneously to comprehensively enhance the performance of CLIP model for the downstream tasks. Give a visual feature $z$ with class $c\in\{1, 2\cdots C\}$ encoded by the visual encoder of CLIP model, the text-based classification loss $\mathcal{L}_{text}$ is calculated as follows:
\begin{equation}
\mathcal{L}_{text}=-log\Big(\frac{exp(sim(z,\tilde{x}^{p}_{c})/\tau)} { {\textstyle \sum_{k=1}^{C}exp(sim(z,\tilde{x}^{p}_{k})/\tau)} }\Big)
\end{equation}
where $sim(\cdot, \cdot)$ denotes the cosine similarity and $\tau$ is a learned temperature parameter of CLIP model. 
To explore the relationships between visual samples, we construct a visual-based classifier, as inspired by Tip-Adapter~\cite{zhang2022tip}. The visual-based classifier leverages a key-value cache model to refine the classification results in CLIP via feature retrieval. For the $C$-class and $K$-shot task, we store all training visual features $X^{cache}\in \mathbb{R}^{CK\times d} $ and the corresponding one-hot label vectors $L\in \mathbb{R}^{CK\times C}$. To obtain task-specific knowledge from the downstream tasks, we learn the visual bias of each class and update the features of all fine-tuning samples as $\widetilde{X}^{cache}$ via the proposed HeGraphAdapter module.
Then, we compute the visual-based classification loss~\cite{zhang2022tip} as follows:
\begin{equation}
\mathcal{L}_{visual}=-log\big(\frac{exp(A_c/\tau)} { {\textstyle \sum_{i=1}^{C}exp(A_i/\tau)} }\big), \; \text{where\;} A=exp\big(sim(z,\widetilde{X}^{cache})-1\big)L
\end{equation}
By adding both losses with a balanced parameter $\lambda$, we obtain the final whole loss function $\mathcal{L}$ as follows:
 \begin{equation}
\mathcal{L}=\mathcal{L}_{text}+\lambda \mathcal{L}_{cache}
\end{equation}

During tuning, we only need to tune the parameters of $W^p$ (Eq.~\ref{SignGCN2}), $W^n$ (Eq.~\ref{SignGCN}) and $W^v$ (Eq.~\ref{SignGCN3}) in our HeGraphAdapter. 
\section{Experiments}\label{sec4}

\subsection{Datasets and Implementation Details}

\textbf{Datasets:} We conduct experiments on few-shot classification and domain generalization tasks to evaluate the effectiveness of our HeGraphAdapter. We select 11 benchmark datasets for few-shot classification tasks based on the previous adapter-style studies~\cite{zhang2022tip,zhou2022learning,li2024graphadapter}, including ImageNet~\cite{deng2009imagenet}, StandfordCars~\cite{krause20133d}, UCF101~\cite{soomro2012ucf101}, Caltech101~\cite{fei2004learning}, Flowers102~\cite{nilsback2008automated}, SUN397~\cite{xiao2010sun}, DTD~\cite{cimpoi2014describing}, EuroSAT~\cite{helber2019eurosat}, FGVCAircraft~\cite{maji2013fine}, OxfordPets~\cite{parkhi2012cats}, and Food101~\cite{bossard2014food}. The selected datasets involve various vision classification tasks, which include remote sensing classification, action recognition, texture classification, and fine-grained classification. Following work GraphAdapter~\cite{li2024graphadapter}, we construct the generalization experiments on the ImageNetV2~\cite{recht2019imagenet}, ImageNet-Sketch~\cite{wang2019learning}, ImageNet-A~\cite{hendrycks2021natural} and ImageNet-R~\cite{hendrycks2021many} datasets.

\textbf{Implementation details:} Our experiments are based on the pre-trained backbone ResNet-50 of CLIP by default. For the positive text prompts, we use ChatGPT to generate multiple text descriptions for each class. The negative text prompt of each class is obtained from a template like ``A photo of no \{class\}". The proposed method is fine-tuned using AdamW optimizer~\cite{diederik2014adam}, which drops with a cosine scheduler. Notably, we adopt a warmup strategy during training, starting with learning rate $1e-5$ to ensure stable training in the first epoch. The learning rate is set to 0.01 for FGVCAircraft dataset and 0.001 for other datasets. Our model is trained 100 epochs for EuroSAT, DTD, and FGVCAircraft datasets, and 30 epochs for other few-shot datasets. 
Due to the difference in the few-shot datasets, we set different values to the fusion weight $\beta$ of meta-path $\Phi^n$, $\alpha$ of meta-path $\Phi^p$, and $\gamma$ of meta-path $\Phi^v$. To preserve more context-based information of positive links, we set the fusion weight of meta-path $n{\rightarrow}p$ to 0.1 during testing. The balance factor $\lambda$ is set to 0.1. All experiments are implemented on a single NVIDIA RTX3090 GPU.

\begin{figure*}[h]
  \centering
  \includegraphics[width=\linewidth]{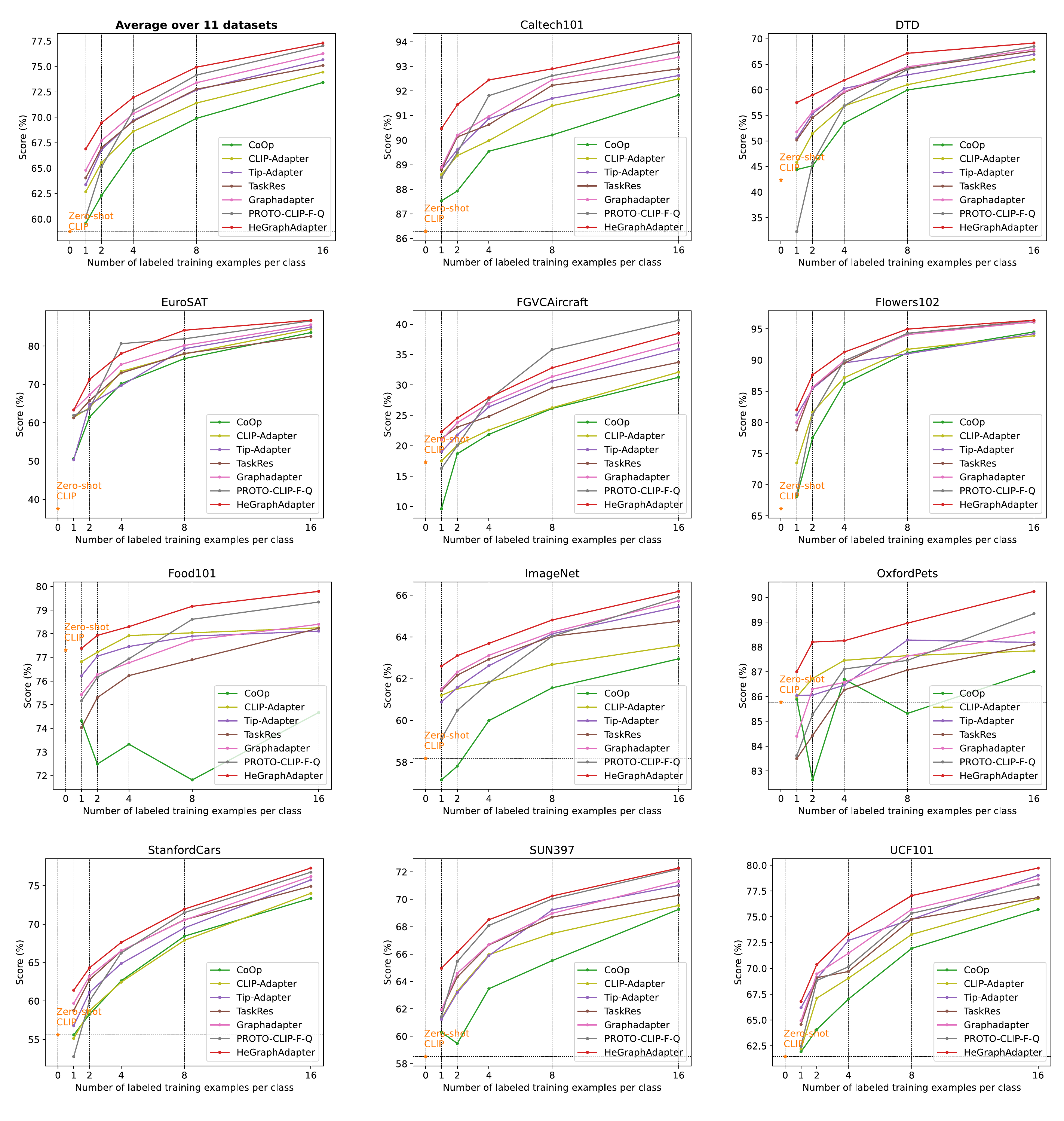}
  \caption{ Comparison results of HeGraphAdapter with the state-of-the-art methods on 11 few-shot datasets.}
\label{results}
\end{figure*}

\subsection{Comparisons with State-of-the-arts}

\textbf{Few-shot learning:} We compare HeGraphAdapter on 11 benchmark datasets with state-of-the-art (SOTA) methods including CLIP-Adapter~\cite{gao2024clip}, CoOp~\cite{zhou2022learning}, TaskRes~\cite{yu2023task}, Tip-Adapter~\cite{zhang2022tip}, GraphAdapter~\cite{li2024graphadapter}, and PROTO-CLIP~\cite{palanisamyproto}. The comparison results are exhibited in Fig.~\ref{results}. The results of state-of-the-art (SOTA) methods have been reported in previous papers and are directly utilized in our experiments. Overall, the average performance of the proposed HeGraphAdapter outperforms previous methods on all few-shot settings. Particularly, our HeGraphAdapter can significantly enhance performance in the 1/2/4-shot setting, with an average performance of 66.89\%, 69.45\%, and 71.93\%. The results exceed the second-best method by 2.09\%, 1.74\%, and 1.28\%. This validates the advantage of learning heterogeneous knowledge across different attribute data for the downstream tasks. Observing the results on the dataset with a large number of classes (1000 classes in ImageNet and 397 classes in SUN397) and a small number of classes (37 classes in OxfordPets), our method demonstrates competitive performance, achieving 66.18\% on ImageNet, 72.28\% on SUN397, and 90.24\% on OxfordPets with 16-shot setting. The results indicate that the proposed HeGraphAdapter can exploit the structure knowledge effectively for both large and small numbers of nodes/classes.

\textbf{Generalization:} As noted by previous method~\cite{zhou2022learning,li2024graphadapter}, training on the specific task carries the risk of learning spurious correlations and reduced generalization ability to unseen domains. To analyze the generalization of HeGraphAdapter, we perform experiments on four unseen commonly-used datasets, including ImageNet-V2~\cite{recht2019imagenet}, ImageNet-Sketch~\cite{wang2019learning}, ImageNet-A~\cite{hendrycks2021natural}, ImageNetR~\cite{hendrycks2021many}. Furthermore, we construct the experiments across different CLIP visual backbones, including ResNet-101~\cite{he2016deep}, ViT-B/32~\cite{dosovitskiy2020image}, and ViT-B/16~\cite{dosovitskiy2020image}. We report the results in Tabel~\ref{gen_res}. The visual-based classifier leverages a key-value cache model from all fine-tuning visual features to refine the classification results, which learn more style information from the downstream tasks. To preserve the generalization ability, we only use the text-based classifier for the generalization experiments, which is denoted as HeGraphAdapter(T) in this section. Compared of Zero-shot CLIP~\cite{radford2021learning}, Linear Probe CLIP~\cite{radford2021learning}, CoOp~\cite{zhou2022learning}, TaskRes~\cite{yu2023task}, and GraphAdapter~\cite{li2024graphadapter}, we achieve superior performance across different backbones, which demonstrates that the heterogeneous graph adapter effectively improves classification accuracy for the downstream tasks while preserving the generalization ability.

\begin{table}[htbp]
\caption{The generalization experiments on various CLIP backbones. The methods are optimized on ImageNet dataset with 16-shot setting denoted as `Source' and tested on cross-domain datasets denoted as `Target'.}
    \begin{tabular}{c|c|cccccc}
    \hline
   \toprule
    \multirow{2}[4]{*}{Method} & \multirow{2}[4]{*}{Backbone} & Source & \multicolumn{5}{c}{Target} \\
\cmidrule{3-8}          &       & ImageNet & -V2   & -Sketch & -A    & -R    & Average \\
    \midrule
    Zero-shot CLIP~\cite{radford2021learning}  &       & 58.18 & 51.34 & 33.32 & 21.65 & 56.00    & 44.10  \bigstrut[t]\\
    Linear Probe CLIP~\cite{radford2021learning} &       & 55.87 & 45.97 & 19.07 & 12.74 & 28.16 & 32.36  \\
    CoOp~\cite{zhou2022learning}  & ResNet-50 & 62.95 & 55.11 & 32.74 & 22.12 & 54.96 & 45.58  \\
    TaskRes~\cite{yu2023task}  &       & 64.75 & 56.47 & \textbf{35.83} & 22.80  & \textbf{60.70} & 48.11  \\
    GraphAdapter~\cite{li2024graphadapter} &       & 65.70  & 56.40  & 34.50  & 21.88 & 58.94 & 47.48  \\
    HeGraphAdapter(T) &       & \textbf{66.06} & \textbf{56.99} & 35.24 & \textbf{23.59} & 59.43 & \textbf{48.26 } \bigstrut[b]\\
    \hline
    Zero-shot CLIP~\cite{radford2021learning}  &       & 61.62 & 54.81 & 38.71 & 28.05 & 64.38 & 49.51  \bigstrut[t]\\
    Linear Probe CLIP~\cite{radford2021learning} &       & 59.75 & 50.05 & 26.80  & 19.44 & 47.19 & 40.65  \\
    CoOp~\cite{zhou2022learning}  &       & 66.60  & 58.66 & 39.08 & 28.89 & 63.00    & 51.25  \\
    TaskRes~\cite{yu2023task}  & ResNet-101 & 67.70  & 59.50  & 41.70  & 29.87 & \textbf{68.07} & 53.37  \\
    GraphAdapter~\cite{li2024graphadapter} &       & 68.23 & 59.60  & 40.83 & 28.77 & 67.13 & 52.91  \\
    HeGraphAdapter(T) &       & \textbf{68.60} & \textbf{59.82} & \textbf{41.88} & \textbf{30.42} & 67.19 & \textbf{53.58 } \bigstrut[b]\\
    \hline
    Zero-shot CLIP~\cite{radford2021learning}  &       & 62.05 & 54.79 & 40.82 & 29.57 & 65.99 & 50.64  \bigstrut[t]\\
    Linear Probe CLIP~\cite{radford2021learning}  &       & 59.58 & 49.73 & 28.06 & 19.67 & 47.20  & 40.85  \\
    CoOp~\cite{zhou2022learning}  &       & 66.85 & 58.08 & 40.44 & 30.62 & 64.45 & 52.09  \\
    TaskRes~\cite{yu2023task}  & ViT-B/32 & 68.20  & 59.20  & 42.50  & \textbf{31.43} & \textbf{69.33} & 54.13  \\
    GraphAdapter~\cite{li2024graphadapter} &       & 68.80  & 59.00    & 41.70  & 29.57 & 68.67 & 53.55  \\
    HeGraphAdapter(T)   &       & \textbf{68.85} & \textbf{59.62} & \textbf{42.77} & 31.42 & 68.61 & \textbf{54.25 } \bigstrut[b]\\
    \hline
    Zero-shot CLIP~\cite{radford2021learning}  &       & 66.73 & 60.83 & 46.15 & 47.77 & 73.96 & 59.09  \bigstrut[t]\\
    Linear Probe CLIP~\cite{radford2021learning} &       & 65.85 & 56.26 & 34.77 & 35.68 & 58.43 & 50.20  \\
    CoOp~\cite{zhou2022learning}  &       & 71.92 & 64.18 & 46.71 & 48.41 & 74.32 & 61.11  \\
    TaskRes~\cite{yu2023task} & ViT-B/16 & 73.07 & 65.30  & 49.13 & 50.37 & \textbf{77.70} & 63.11  \\
    GraphAdapter~\cite{li2024graphadapter} &       & 73.68 & \textbf{65.57} & 48.57 & 49.23 & 77.20  & 62.85  \\
    HeGraphAdapter(T)   &       & \textbf{73.82} & 65.39 & \textbf{49.31} & \textbf{50.78} & 77.19 & \textbf{63.30 } \bigstrut[b]\\
    \hline
    \end{tabular}%
    \label{gen_res}
\end{table}%

\subsection{ Ablation Studies}
\textbf{Effectiveness of different components:} 
To verify the effectiveness of the heterogeneous graph adapter, we conduct several experiments on 11 few-shot datasets with 16-shot setting. The `Base' method only utilizes the similarity matching between positive text prompts and test images. HeGraphAdapter(T) only uses the text-based classifier for experiments. HeGraphAdapter(T-N) only considers negative text node aggregation on the heterogeneous graph adapter. HeGraphAdapter(T-P) only considers the relations between positive text prompts and visual features. As shown in Table~\ref{ablation}, the `Base' method achieves an average performance of 60.19\%. After introducing the negative text prompts and only considering negative node aggregation on the heterogeneous graph adapter as HeGraphAdapter(T-N), we get an improvement of 8.17\%. By considering the negative and positive text prompts simultaneously as HeGraphAdapter(T), the performance improves by 17.07\% compared to the `Base' method. This demonstrates that positive and negative text prompts are all necessary to enhance the performance on the few-shot classification tasks. HeGraphAdapter method achieves an average performance of 77.30\% and improves by 0.04\% compared to HeGraphAdapter(T), which indicates that refining both the text-based and visual-based classifier simultaneously is effective for comprehensively enhancing the performance of CLIP.

\begin{table*}[htbp]
\setlength\tabcolsep{3pt}
\caption{Ablation study of HeGraphAdapter. `Base' only utilizes the similarity matching between positive text prompts and test images. HeGraphAdapter(T-N) only considers negative text node aggregation on the heterogeneous graph adapter. HeGraphAdapter(T-P) only considers the relations between positive text prompts and visual features. HeGraphAdapter(T) only uses the text-based classifier for experiments. HeGraphAdapter utilizes both text-based and visual-based classifiers simultaneously. }
 \resizebox{1\textwidth}{!}{
 \begin{tabular}{c|cccccccccccc}
    \hline
    \rotatebox{90}{Method} & \rotatebox{90}{ImageNet} & \rotatebox{90}{Caltech101} & \rotatebox{90}{OxfordPets} & \rotatebox{90}{StanfordCars} & \rotatebox{90}{Flowers102} & \rotatebox{90}{Food101} & \rotatebox{90}{FGVCAircraft} & \rotatebox{90}{SUN397} & \rotatebox{90}{DTD}   & \rotatebox{90}{EuroSAT} & \rotatebox{90}{UCF101} & \rotatebox{90}{Average} \bigstrut\\
    \hline
    CoOp~\cite{zhou2022learning}  & 62.95 & 91.83 & 87.01 & 73.36 & 94.51 & 74.67 & 31.26 & 69.26 & 63.58 & 83.53 & 75.71 & 73.42  \\
    TaskRes~\cite{yu2023task} & 64.75 & 92.90  & 88.10 & 74.93 & 96.10  & 78.23 & 33.73 & 70.30  & 67.57 & 82.57 & 76.87 & 75.10  \\
    Tip-Adapter~\cite{zhang2022tip} & 65.44 & 92.63 & 88.18 & 75.75 & 94.23 & 78.11 & 35.86 & 71.00    & 66.94 & 84.94 & 79.03 & 75.65  \\
    CLIP-Adapter~\cite{gao2024clip} & 63.59 & 92.49 & 87.84 & 74.01 & 93.90  & 78.25 & 32.10  & 69.55 & 65.96 & 84.43 & 76.76 & 74.44 \\
    GraphAdapter~\cite{li2024graphadapter} & 65.72 & 93.37 & 88.59 & 76.20  & 96.13 & 78.40  & 36.93 & 71.30  & 67.90  & 85.55 & 78.67 & 76.25  \\
    PROTO-CLIP-F-Q~\cite{palanisamyproto} & 65.91 & 93.59 & 89.34 & 76.76 & 96.35 & 79.34 & 40.65 & 72.19 & 68.50  & 86.59 & 78.11 & 77.03 \\
    \hline
    Base  & 61.40  & 89.45 & 84.46 & 57.27 & 65.29 & 76.77 & 19.29 & 62.47 & 48.46 & 39.54 & 57.73 & 60.19  \\
    HeGraphAdapter(T-N)     & 63.13 & 91.24 & 87.16 & 68.37 & 82.10  & 77.69 & 25.92 & 66.72 & 56.03 & 64.01 & 69.55 & 68.36  \\
    HeGraphAdapter(T-P) & 65.98 & 93.83 & 89.97 & 77.28 & 96.26 & 79.50  & 38.04 & 72.20  & 68.62 & 86.70  & 79.46 & 77.08  \\
    HeGraphAdapter(T) & 66.06 & 93.91 & 90.16 & \textbf{77.34} & 96.31 & 79.78 & 38.31 & 72.27 & \textbf{69.27} & \textbf{86.77} & 79.65 & 77.26  \\
    HeGraphAdapter & \textbf{66.18} & \textbf{93.96} & \textbf{90.24} & 77.30  & \textbf{96.39} & \textbf{79.79} & \textbf{38.49} & \textbf{72.28} & 69.15 & 86.75 & \textbf{79.73} & \textbf{77.30 } \\
    \hline
    \end{tabular}}
    \label{ablation}%
\end{table*}%

\textbf{Sensitivity analysis of hyper-parameters:}
To verify the sensitivity of the fused weight $\alpha$ in Eq.~\ref{alpha} via meta-path $\Phi^p=\left\{ p{\rightarrow}p, v{\rightarrow}p, n{\rightarrow}p\right \}$, we conduct several experiments that only utilize the text-based classifier, and the results are shown in Table~\ref{sen}. The weight of meta-path $v{\rightarrow}p$ is larger than the weight of meta-path $p{\rightarrow}p$, which demonstrates that constructing the heterogeneous relations between text and visual data is crucial for enhancing the distinctiveness of CLIP model. Otherwise, we model the dissimilarity relations between negative and positive text nodes based on the meta-path $n{\rightarrow}p$. During fine-tuning, we assign a larger weight to $\alpha^{n{\rightarrow}p}$ to promote the network to focus more on the negative text nodes. However, during testing, we reduce this weight to preserve more information, as the negative text nodes are connected via negative links. We also conduct an experiment where all fused weights of the meta-paths are learnable. We achieve a performance of 65.68\%, which is lower than the performance of the fixed coefficients. The drop in performance can be attributed to the difficulty in optimizing the coefficients with the limited data available in few-shot scenarios, leading to sub-optimal learning.

\begin{table*}[htbp]
  \centering
  \caption{Sensitivity analysis of hyper-parameters $\alpha$ on ImageNet dataset with 16-shot setting.}
  \resizebox{1\textwidth}{!}{
    \begin{tabular}{ccccccccccc|c}
    \toprule
    $\alpha^{ p{\rightarrow}p}(\alpha^{v{\rightarrow}p}=0.24,\alpha^{n{\rightarrow}p}=0.2)$ & 0     & 0.02  & \textbf{0.06}  & 0.1   & 0.14  & 0.18  & 0.22  & 0.26  & 0.3   & 0.4   & \multicolumn{1}{l}{learn} \\
    ImageNet & 65.96    & 66.00 & 66.06 & 66.05 & 66.03  & 65.95 & 65.89 & 65.84 & 65.80 & 65.64 & \multirow{7}[7]{*}{65.68} \\
\cmidrule{1-11}    $\alpha^{v{\rightarrow}p}(\alpha^{p{\rightarrow}p}=0.06,\alpha^{n{\rightarrow}p}=0.2)$ & 0     & 0.06  & 0.12  & 0.18  & \textbf{0.24}  & \textbf{0.3}   & 0.36  & 0.42  & 0.5   & 0.6   &  \\
    ImageNet & 64.54 & 65.28 & 65.65 & 65.96 & 66.06 & 66.06 & 65.95 & 65.89 & 65.85 & 65.64 &  \\
\cmidrule{1-11}    $\alpha^{n{\rightarrow}p} (\alpha^{p{\rightarrow}p}=0.06;\alpha^{n{\rightarrow}p}=0.24)$ & 0     & 0.05  & 0.1   & 0.15  & \textbf{0.2}   & \textbf{0.25}  & 0.3   & 0.4   & 0.5   & 0.6   &  \\
    ImageNet & 65.73 & 65.75 & 65.88 & 65.92 & 66.06 & 66.06 & 65.99 & 65.97 & 65.91 & 65.84  &  \\
\cmidrule{1-11}    $test: \alpha^{n{\rightarrow}p}$ & 0     & 0.05  & \textbf{0.1}   & 0.15  & 0.2   & 0.25  & 0.3   & 0.4   & 0.5   & 0.6   &  \\
    ImageNet & 66.00    & 66.05 & 66.06 & 65.70 & 65.70 & 65.43 & 65.18 & 64.43 & 63.48 & 61.88 &  \\
    \bottomrule
    \end{tabular}%
  \label{sen}%
  }
\end{table*}%

\textbf{Effects of different backbones:} We construct extensive experiments on ImageNet dataset with 16-shot setting to evaluate the effectiveness of the proposed method on different CLIP backbones. The results are displayed in Table~\ref{backbone}. The CLIP backbone includes ResNet-50, ResNet-101, ViT-B/32, and
ViT-B/16. Our method achieves performances of 66.18\% with ResNet-50, 68.66\% with ResNet-101, 69.20\% with ViT-B/32, and 74.02\% with ViT-B/16, outperforming the second-best method by 0.46\%, 0.43\%, 0.40\%, and 0.34\%, respectively. This demonstrates that our HeGraphAdapter is effective across various visual backbones.

\begin{table}[htbp]
  \centering
  \caption{Results of different CLIP backbones on ImageNet dataset with 16-shot setting.}
    \begin{tabular}{c|cccc}
    \toprule
    Method & ResNet-50 & ResNet-101 & ViT-B/32 & ViT-B/16 \\
    \midrule
    Zero-shot CLIP~\cite{radford2021learning}  & 58.18 & 61.62 & 62.05 & 66.73 \\
    Linear Probe CLIP~\cite{radford2021learning} & 55.87 & 59.75 & 59.58 & 65.85 \\
    CoOp~\cite{zhou2022learning}  & 62.95 & 66.60  & 66.85 & 71.92 \\
    TaskRes~\cite{yu2023task}  & 64.75 & 67.70  & 68.20  & 73.07 \\
    Tip-Adapter-F~\cite{zhang2022tip} & 65.51 & 68.56  & 68.65  & 73.69 \\ 
    GraphAdapter~\cite{li2024graphadapter} & 65.72 & 68.23 & 68.80  & 73.68 \\
    HeGraphAdapter   & \textbf{66.18} & \textbf{68.66} & \textbf{69.20} & \textbf{74.02} \\
    \bottomrule
    \end{tabular}%
  \label{backbone}%
\end{table}%

\begin{figure*}[h]
  \centering
  \includegraphics[width=0.6\linewidth]{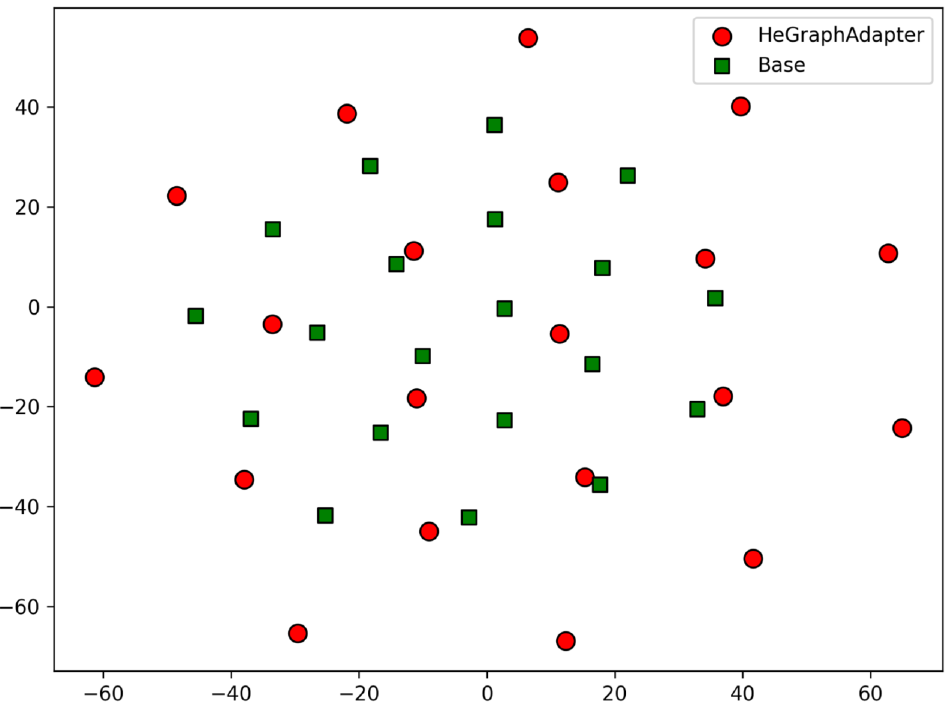}
  \caption{2D t-SNE visualization of positive textual features before and after HeGraphAdapter. Each node corresponds to a class.}
\label{t-sne}
\end{figure*}

\textbf{Visualization:}
To demonstrate that the heterogeneous graph adapter can capture discriminative textual features, we leverage the 2D t-SNE~\cite{van2008visualizing} to illustrate the distribution of positive text nodes as shown in Fig.~\ref{t-sne}, where each node represents a class. We randomly sampled 20 classes from the fine-grained Food101 dataset to better present the effectiveness of  HeGraphAdapter. The comparison results between the `Base' model and our `HeGraphAdapter' demonstrate that the proposed heterogeneous graph adapter effectively models the relationships between different modalities, achieving larger inter-class distances and thus can enhance the classification ability of adapter-style tuning.

\textbf{Complexity Analysis:}
In Table~\ref{time}, we show the fine-tuning time and the number of tunable parameters for HeGraphAdapter on ImageNet dataset with 16-shot setting. The results of HeGraphAdapter are achieved on a single NVIDIA RTX3090 GPU. The comparison results are sourced from their papers and we use them directly in our experiments. From the results, we can observe that HeGraphAdapter requires 33.95 seconds to fine-tune one epoch, which is longer than the GraphAdapter~\cite{li2024graphadapter} method at 10.66 seconds, but HeGraphAdapter requires fewer fine-tuning epochs. In HeGraphAdapter, we only need to tune the parameters of $W^p$, $W^n$, and $W^v$. The number of tunable parameters is 10.367M. Although our HeGraphAdapter requires a larger number of tunable parameters, its performance surpasses the second-best method by 0.46\%, which achieves a balance between efficiency and performance.

\begin{table*}[htbp]
  \centering
  \caption{Comparison results of fine-tuning time and complexity on ImageNet dataset with 16-shot setting. }
  \resizebox{1\textwidth}{!}{
    \begin{tabular}{c|c|c|c|c}
    \hline
    Method & Training(One epoch)(s) & Epochs & Tunable Parameters(M) & Performance \\
    \hline
    CoOp~\cite{zhou2022learning}  & 40.91 & 200   & 0.008 & 62.95 \\
    CLIP-Adapter~\cite{gao2024clip} & 45.71 & 200   & 0.524 & 63.59 \\
    Tip-Adapter-F~\cite{zhang2022tip} & 12.36 & 20    & 16.380 & 65.51 \\
    GraphAdapter~\cite{li2024graphadapter} & 23.29 & 100   & 4.145 & 65.72 \\
    HeGraphAdapter & 33.95 & 30    & 10.367 & 66.18 \\
    \hline
    \end{tabular}%
    }
  \label{time}%
\end{table*}%

\section{Conclusion}

This paper proposes a new heterogeneous graph adapter for fine-tuning VLMs. We first construct a heterogeneous graph to comprehensively model the relationships of intra-modality, inter-modality and inter-class, and then develop a heterogeneous graph neural network to extract multi-modality structure knowledge for adapting both visual and textual features to the downstream tasks. Moreover, we utilize the specific-task textual and visual features to construct both classifiers and enhance the performance of CLIP for the downstream tasks. 
Extension experiments on eleven benchmark datasets demonstrate the effectiveness of the proposed method. 
We think heterogeneous graph learning
provides an effective model for the general multi-modality representation problem. 
In the future, we will explore heterogeneous graph learning to address the fine-tuning problem for some other pre-trained multi-modality models. 
\bibliography{article}

\end{document}